\documentclass{article} 
\usepackage{iclr2025_conference,times}

\usepackage{amsmath,amsfonts,bm}
\usepackage{dsfont}
\usepackage{xspace}

\def\eqref#1{equation~\ref{#1}}

\def\1{\bm{1}}

\DeclareMathAlphabet{\mathsfit}{\encodingdefault}{\sfdefault}{m}{sl}
\SetMathAlphabet{\mathsfit}{bold}{\encodingdefault}{\sfdefault}{bx}{n}

\newcommand{\pycm}{PyCM\xspace}

\usepackage{dsfont}
\usepackage{tikz}
\usepackage{fancyvrb}
\usepackage{hyperref}
\usepackage{url}
\usepackage{multirow}
\usepackage{booktabs}
\usepackage{rotating, graphicx}
\usepackage{tabularx, makecell}
\usepackage{pythonhighlight}
\usepackage{pgfplots}
\usepackage{subcaption}
\usepgfplotslibrary{colormaps,colorbrewer}
\pgfplotsset{compat=1.18}

\title{Comparing Classifiers: A Case Study Using \pycm}

\iclrfinalcopy

\author{
  Sadra Sabouri\thanks{Equal contribution} \textsuperscript{,1,2}, 
  Alireza Zolanvari\footnotemark[1] \textsuperscript{,1} \& 
  Sepand Haghighi\footnotemark[1] \textsuperscript{,1} \\
  \textsuperscript{1}Open Science Laboratory \\
  \textsuperscript{2}Department of Computer Science, University of Southern California \\
  \texttt{\{sadra,alireza,sepand\}@openscilab.com}
}

\begin{document}

\maketitle

\begin{abstract}
Selecting an optimal classification model requires a robust and comprehensive understanding of the performance of the model. This paper provides a tutorial on the \pycm library, demonstrating its utility in conducting deep-dive evaluations of multi-class classifiers. By examining two different case scenarios, we illustrate how the choice of evaluation metrics can fundamentally shift the interpretation of a model’s efficacy. Our findings emphasize that a multi-dimensional evaluation framework is essential for uncovering small but important differences in model performance. However, standard metrics may miss these subtle performance trade-offs.
\end{abstract}

\section{Introduction}
Evaluating the performance of machine learning classifiers is a crucial step in building reliable predictive models, particularly in applications where errors can have severe consequences. In medical diagnosis, for example, the stakes are particularly high. An inaccurate cancer detection system can miss patients who actually have cancer, delay their treatment and potentially worsen their prognosis~\cite{long2011causes}, while a false positive can subject a healthy patient to unnecessary stress and invasive high-risk procedures~\cite{carnahan2023false}.

Despite these complexities, many studies still rely on accuracy as their primary benchmark. Although it is easy to calculate, accuracy provides an incomplete and sometimes misleading view of performance. This becomes a major issue in imbalanced datasets where one class is much more frequent than others~\cite{chawla2010data,zolanvari2022literature}.
Consider Huntington’s disease, which affects roughly 1 in 10,000 people~\cite{fields2013genetic}. A trivial classifier that labels all patients as healthy would achieve 99.99\% accuracy. On paper, it looks nearly perfect, yet it is entirely useless for actual clinical work~\cite{ghanem2023limitations}.

To address these limitations, researchers have developed a variety of alternative metrics that capture different aspects of classifiers' performance. For example, the F1-score that balances precision and recall~\cite{sammut2011encyclopedia}, and other metrics such as sensitivity~\cite{yerushalmy1947statistical}, specificity~\cite{yerushalmy1947statistical}, the area under the ROC curve (AUC)~\cite{swets2014signal}, and Cohen’s kappa~\cite{mchugh2012interrater} each highlight different trade-offs depending on the goals of the project.

Choosing the right metric and interpreting it correctly is crucial for performance evaluation and decision-making. Poor metric selection can lead to the choice of suboptimal models, flawed decisions, and ultimately the deployment of systems that fail or even cause unintended harm and biases in real-world settings~\cite{federspiel2023threats}. This paper presents two realistic but hypothetical data-driven decision-making scenarios that provide a practical guide on understanding different evaluation metrics, choosing metrics suited to specific contexts, and interpreting what they indicate about classifier performance.

In this tutorial, we begin by positioning \pycm~\cite{haghighi2018pycm} framework within the standard machine learning pipeline, highlighting its role as a critical component for automated statistical analysis and transparent, well-justified model selection (\S \ref{sec:workflow}).
Following this, we establish the core definitions and notations used throughout the text (\S \ref{sec:prelims}). We then examine various methodologies for model quality comparison (\S \ref{sec:compare}) and present real-world scenarios where evaluating and comparing classifiers becomes particularly complex, demonstrating how \pycm bridges these gaps by providing accessible, robust comparison methods (\S \ref{sec:case_study}).

\section{\pycm in Machine Learning Pipeline}
\label{sec:workflow}
\begin{figure}[tbh]
    \centering
    \includegraphics[ width=0.5\linewidth]{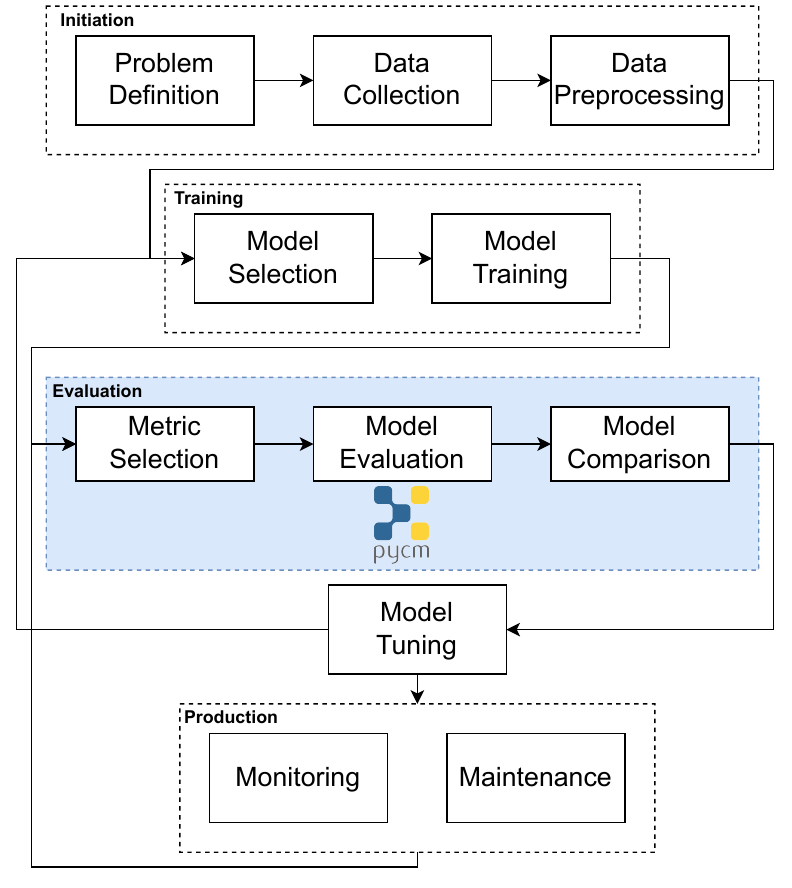}
    \caption{\pycm in the Machine Learning Workflow. The workflow starts with \textbf{Initiation} (Problem Definition, Data Collection, and Preprocessing), followed by \textbf{Training} (Model Selection and Training). \textbf{Evaluation}, where \pycm is used, includes Metric Selection and Model Comparison. Results from the evaluation guide Model Tuning. The final phase is \textbf{Production}, which includes Monitoring and Maintenance to ensure long-term reliability.}
    \label{fig:pycm-in-ml-workflow}
\end{figure}

\pycm is an open-source Python library that is designed to fill the gap for evaluating machine learning models specifically for multi-class classification tasks~\cite{haghighi2018pycm}.
In a supervised machine learning workflow, it serves as a systematic evaluation and reporting framework for data scientists and machine learning engineers to accurately assess and compare the performance of various classifiers. Figure \ref{fig:pycm-in-ml-workflow} illustrates such a workflow and highlights the stage where \pycm is utilized.

The typical lifecycle of a machine learning model encompasses the following key stages:

\textbf{Initiation.}
The first step is to define the prediction task. This can be regression (predicting a number) or classification (assigning a category). We need a dataset with enough examples of input features and expected outputs. Then, the data would be prepared and formatted so the model can use it for training.

\textbf{Training.}
After data preparation, we choose a model. Model choice depends on the problem, the dataset size, and the type of task. Heuristics or simple rules can help select a suitable model~\footnote{For example: \url{https://scikit-learn.org/stable/machine_learning_map.html}}. Then, the model is trained to learn patterns from the data.

\textbf{Evaluation.}
After training, it is critical to evaluate how well the model performs. Choosing the right metrics is important because they should match the goals of the task. Metrics convert complex model behaviors into clear numerical values, allowing us to compare different models objectively.

\pycm is designed specifically for this phase. It provides over 150 metrics to evaluate performance across various data distributions. In addition, the \texttt{Compare} interface ranks multiple models, simplifying the selection process.

The evaluation typically informs the model design. If the evaluation shows that the model does not meet the required standards, the workflow returns to the training phase for adjustments. This cycle continues until the model meets the desired performance.

\textbf{Production}
When the model passes the evaluation phase, it is ready to be deployed to production. In this final stage, continuous monitoring is essential. It should be continuously evaluated to check how the model generalizes to new real-world data and identify any unexpected errors. Data collected during production would also be ingested back into the previous stages for further updates and improvements in the effective handling of real-world interactions.

\section{Definitions}
\label{sec:prelims}
To build an understanding of the different evaluation methods, we need to establish the formal notation and definitions used to quantify classifier performance. These core concepts provide the mathematical foundation for the comparative analysis and metric interpretations presented in the following chapters.

\subsection{Notation}
\label{subsec:notation}

\textbf{Classes ($\mathcal{C}$)} represent the set of all possible categories that the classification model can predict. The choice of classes depends entirely on the application domain. For example, for a medical diagnosis system identifying diseases, you might have $\mathcal{C} = \{\text{Healthy}, \text{Flu}, \text{COVID}\}$.

\textbf{Predictions and Actual Labels} are two parallel lists of length $n$ such that:
\begin{itemize}
    \item \fbox{\Verb|act|} contains the actual categories (labels) that each data point belongs to,
    \item \fbox{\Verb|pred|} contains the labels that the classification model predicted for each data point.
\end{itemize}
For example, for a diagnosis system processing $n=5$ patient cases, the input vectors may appear as follows:
\begin{verbatim}
pred = ['Flu',   'Healthy', 'COVID', 'Flu', 'Healthy']
act  = ['COVID', 'Healthy', 'Flu',   'Flu', 'COVID']
\end{verbatim}
In this example, the model correctly identifies two instances: one for \verb|Healthy| and one for \verb|Flu|, the second and the last diagnosis. The remaining cases illustrate specific classification errors: the first case misidentifies COVID as flu, the third case falsely classifies flu as COVID, and the final case predicts a healthy outcome for a patient with COVID.

\textbf{Indexed Metrics ($F_i$)} denotes the value of a specific performance measure calculated for a single class $i \in \mathcal{C}$. Subscript notation enables the analysis of class-specific behavior rather than aggregate averages. For instance, the class-wise accuracy for the Flu category is $ACC_{\text{Flu}} = \frac{1}{5}$, reflecting the single instance where both \verb|pred| and \verb|act| identify the class correctly. In this example, the class-based accuracies for other classes are as follows: $ACC_{\text{Healthy}} = \frac{1}{5}$ and $ACC_{\text{COVID}} = 0$.

\textbf{Confusion Matrix} is a tabular representation of the relationship between predicted and actual labels, quantifying how frequently a model confuses distinct categories. We use $CM$ to refer to the mathematical concept and \fbox{\Verb|cm|} to denote the specific \fbox{\Verb|ConfusionMatrix(act, pred)|} object within the \pycm library. The entry at row $i$ and column $j$ represents the number of instances where an actual class $i$ was predicted as class $j$. For example, $CM_{\text{COVID, Healthy}} = 1$ indicates a single case where a patient with COVID was misdiagnosed as healthy. For the diagnosis classifier example, the respective confusion matrix is shown in Table~\ref{tab:diag-example-cm}.
{\renewcommand{\arraystretch}{1.2}
\begin{table}[h]
    \centering
    \begin{tabular}{c|p{3em}|p{3em}|p{3em}|p{3em}|}
         \multicolumn{2}{c}{} & \multicolumn{3}{c}{Predicted} \\
          \cline{3-5}
         \multicolumn{2}{c|}{} & Healthy & Flu & COVID \\
          \cline{2-5}
         \multirow{3}{*}{\rotatebox[origin=c]{90}{Actual}} & Healthy & 1 & 0 & 0\\
          \cline{2-5}
         & Flu & 0 & 1 & 1\\
          \cline{2-5}
         & COVID & 1 & 1 & 0\\
          \cline{2-5}
    \end{tabular}
    \caption{A confusion matrix representing the performance of a diagnosis classifier system.}
    \label{tab:diag-example-cm}
\end{table}}

\textbf{The Indicator Function ($\mathds{1}\{X\}$)} is a mathematical operator that maps a logical statement $X$ to the set $\{0, 1\}$, facilitating the formalization of counting operations.
\begin{equation}
\mathds{1}\{X\} = \begin{cases}1 & \text{if } X \text{ is true} \\0 & \text{if } X \text{ is false}\end{cases}
\end{equation}
For example, this function allows for the formal definition of overall accuracy ($ACC_{\text{overall}}$) as the proportion of matching elements between the prediction and actual vectors:

\textbf{Code Convention.} We use \fbox{\Verb|monospaced font|} to distinguish Python objects, functions, and variable names from theoretical notation. This distinction clarifies the transition between mathematical concepts and their implementation, such as using \fbox{\Verb|cm.TP|} to access the True Positive counts stored within a \pycm object.

\subsection{Confusion Matrix Elements}
\label{subsec:fund-metrics}
Although confusion matrices generalize to any number of classes, binary classification remains the basic standard for establishing core terminology. In this context, the set of classes is defined as $\mathcal{C} = \{0, 1\}$, where $0$ and $1$ represent the negative and positive classes, respectively. Table~\ref{tab:binar-confusion-matrix} illustrates this structure and provides the basis for the following definitions.

{
\renewcommand{\arraystretch}{2}
\begin{table}[th]
    \centering
    \begin{tabular}{c|c|c|c|}
         \multicolumn{2}{c}{} & \multicolumn{2}{c}{Predicted} \\
          \cline{3-4}
         \multicolumn{2}{c|}{} & 0 & 1 \\
          \cline{2-4}
         \multirow{2}{*}{\rotatebox[origin=c]{90}{Actual}} & \rotatebox[origin=c]{90}{0} & $TN$ & $FP$ \\
          \cline{2-4}
         & \rotatebox[origin=c]{90}{1} & $FN$ & $TP$ \\
          \cline{2-4}
    \end{tabular}
    \caption{Binary confusion matrix with standard notation}
    \label{tab:binar-confusion-matrix}
\end{table}
}

\textbf{True Negative (TN)} is the number of samples correctly classified as the negative class ($0$)~\cite{sammut2011encyclopedia}:
\begin{equation}
    TN=\sum_{i=1}^n\mathds{1}\{\text{pred}_i=0, \text{act}_i=0\}
\end{equation}
In \pycm, this is accessible through \fbox{\Verb|cm.TN|}.

\textbf{False Positive (FP)}, or ``Type I error'', denotes the number of negative samples ($0$) incorrectly predicted as positive ($1$)~\cite{sammut2011encyclopedia}:
\begin{equation}
    FP=\sum_{i=1}^n\mathds{1}\{\text{pred}_i=1, \text{act}_i=0\}
\end{equation}
In \pycm, this is accessible through \fbox{\Verb|cm.FP|}.

\textbf{False Negative (FN)}, or ``Type II error'', denotes the number of positive samples ($1$) incorrectly predicted as negative ($0$)~\cite{sammut2011encyclopedia}:
\begin{equation}
    FN=\sum_{i=1}^n\mathds{1}\{\text{pred}_i=0, \text{act}_i=1\}
\end{equation}
In \pycm, this is accessible through \fbox{\Verb|cm.FN|}.

\textbf{True Positive (TP)} is the number of samples correctly classified as positive class ($1$)~\cite{sammut2011encyclopedia}:
\begin{equation}
    TP=\sum_{i=1}^n\mathds{1}\{\text{pred}_i=1, \text{act}_i=1\}    
\end{equation}
In \pycm, this is accessible through \fbox{\Verb|cm.TP|}.

\subsection{Evaluation Metrics}
\label{subsec:basic-metrics}
Based on the elements defined in \S\ref{subsec:fund-metrics} we now present list a set of evaluation metrics.
Each of the following metrics focuses on and captures a specific aspect of the performance of the classifier.

\textbf{ACC (Accuracy)} calculates the ratio of correct predictions to the total number of instances~\cite{sammut2011encyclopedia}. While intuitive, accuracy can be misleading in imbalanced datasets, where a high score may simply reflect the distribution of the majority class.
\begin{equation}
    ACC = \frac{TN+TP}{n}
\end{equation}

In \pycm, \fbox{\Verb|cm.ACC|} returns the class-based accuracy, \fbox{\Verb|cm.Overall\_ACC|} provides the overall accuracy for all classes, and \fbox{\Verb|cm.ACC\_Macro|} returns the macro accuracy~\cite{sammut2011encyclopedia}.

\textbf{TPR (True Positive Rate)}, also known as sensitivity or recall, measures the proportion of actual positive instances correctly identified by the model~\cite{sammut2011encyclopedia}. High TPR indicates a model's efficacy in minimizing missed cases (False Negatives).

\begin{equation}
\label{eq-TPR}
    TPR = \frac{TP}{FN+TP}
\end{equation}
In \pycm, class-based TPR is available via \fbox{\Verb|cm.TPR|}. The macro and micro TPR values~\cite{sammut2011encyclopedia} are also accessible through \fbox{\Verb|cm.TPR\_Macro|} and \fbox{\Verb|cm.TPR\_Micro|}, respectively.

\textbf{TNR (True Negative Rate)}, or specificity, measures the proportion of actual negative instances correctly predicted~\cite{sammut2011encyclopedia}. This metric evaluates a model's ability to accurately exclude negative cases.

\begin{equation}
    TNR = \frac{TN}{TN+FP}
\end{equation}
In \pycm, class-based TNR is available via \fbox{\Verb|cm.TNR|}. The macro and micro TNR values~\cite{sammut2011encyclopedia} are also accessible through \fbox{\Verb|cm.TNR\_Macro|} and \fbox{\Verb|cm.TNR\_Micro|}, respectively.

\textbf{PPV (Positive Predictive Value)}, or precision, is a measure that quantifies the proportion of true positive predictions (correctly identified positive instances) among all instances that were predicted as positive by the model~\cite{sammut2011encyclopedia}. It is an important metric when the cost of falsely predicting a negative instance as positive (Type I error) is high. A high PPV indicates that the model is accurate in its positive predictions, while a low PPV suggests that the model is making a significant number of false positive predictions.

\begin{equation}
    PPV = \frac{TP}{TP + FP}
\end{equation}
\fbox{\Verb|cm.PPV|}, in \pycm would give you class-based PPV. For multi-class classification, if you wanted to derive the macro and micro PPV~\cite{sammut2011encyclopedia}, use \fbox{\Verb|cm.PPV\_Macro|} and \fbox{\Verb|cm.PPV\_Micro|} respectively.

\textbf{F1 Score} is a single metric that combines precision and recall into a unified measure of a model's accuracy in binary classification problems. F1 Score is widely used in evaluating classification models, especially when dealing with imbalanced datasets or when there is a need to balance trade-offs between precision and recall~\cite{sammut2011encyclopedia}.

\begin{equation}
\begin{aligned}
    \text{F}_1 = \frac{2 \times PPV \times TPR}{PPV + TPR}
\end{aligned}
\end{equation}
The class-based $F_{1}$ is available in \pycm using \fbox{\Verb|cm.F1|}.
For $F_1$ macro, micro use \fbox{\Verb|cm.F1\_Macro|} and \fbox{\Verb|cm.F1\_Micro|} respectively.

\subsection{Curves}
\label{subsec:curves}
Certain classification models output a probability distribution over classes rather than discrete labels, indicating the likelihood of association with each class. In these instances, final class assignments depend on a defined threshold $t$ that converts these probabilistic values into discrete ones. Selecting an optimal threshold involves balancing competing metrics, such as precision and recall. Visualizing these trade-offs through performance curves allows for a comprehensive evaluation of model behavior across all possible threshold values.

Let $f, g: [0, 1] \to [0, 1]$ represent two performance metrics that vary as a function of the threshold~$t$. We define the curve for $f(t)$ and $g(t)$ as the set of coordinate pairs:
\begin{equation}
\label{eq:curve}
    \text{Curve}_{f,g} = \{(f(t), g(t)) \mid t \in [0, 1]\}
\end{equation}

\begin{figure}[htbp]
    \centering
    \begin{subfigure}{0.48\textwidth}
        \centering
        \includegraphics[width=\linewidth]{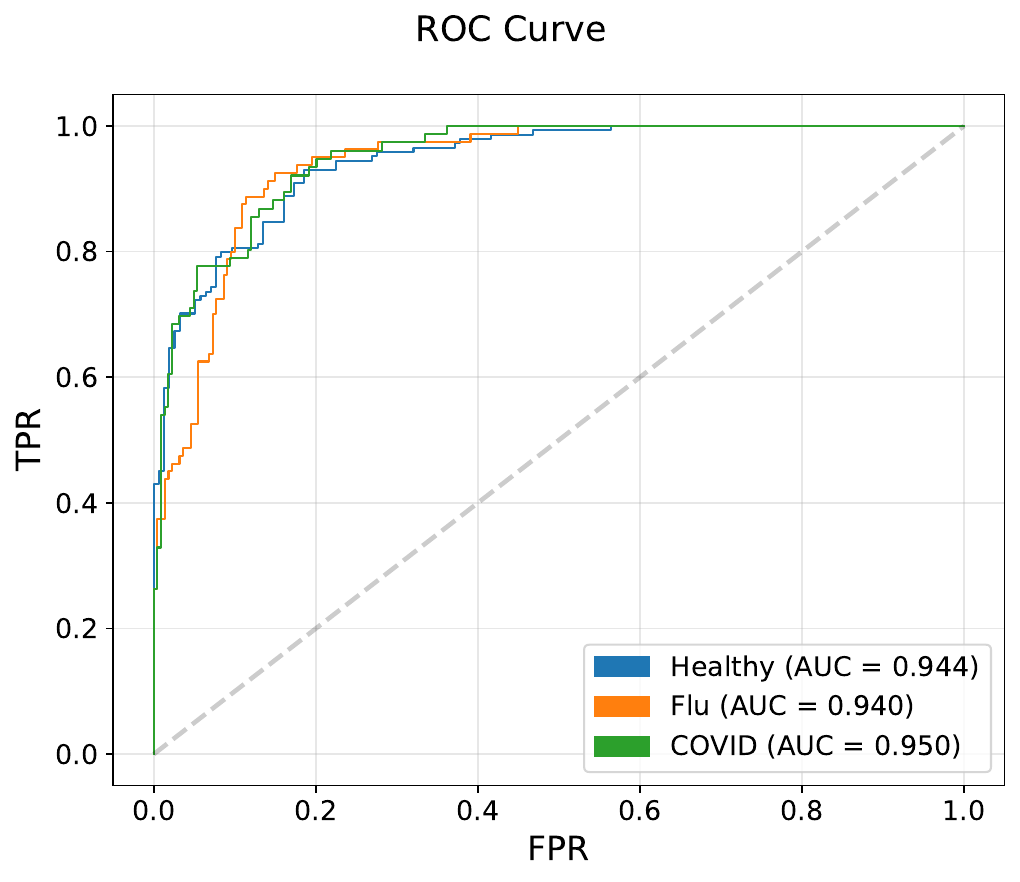}
        \caption{}
        \label{fig:curves:roc}
    \end{subfigure}
    \hfill
    \begin{subfigure}{0.48\textwidth}
        \centering
        \includegraphics[width=\linewidth]{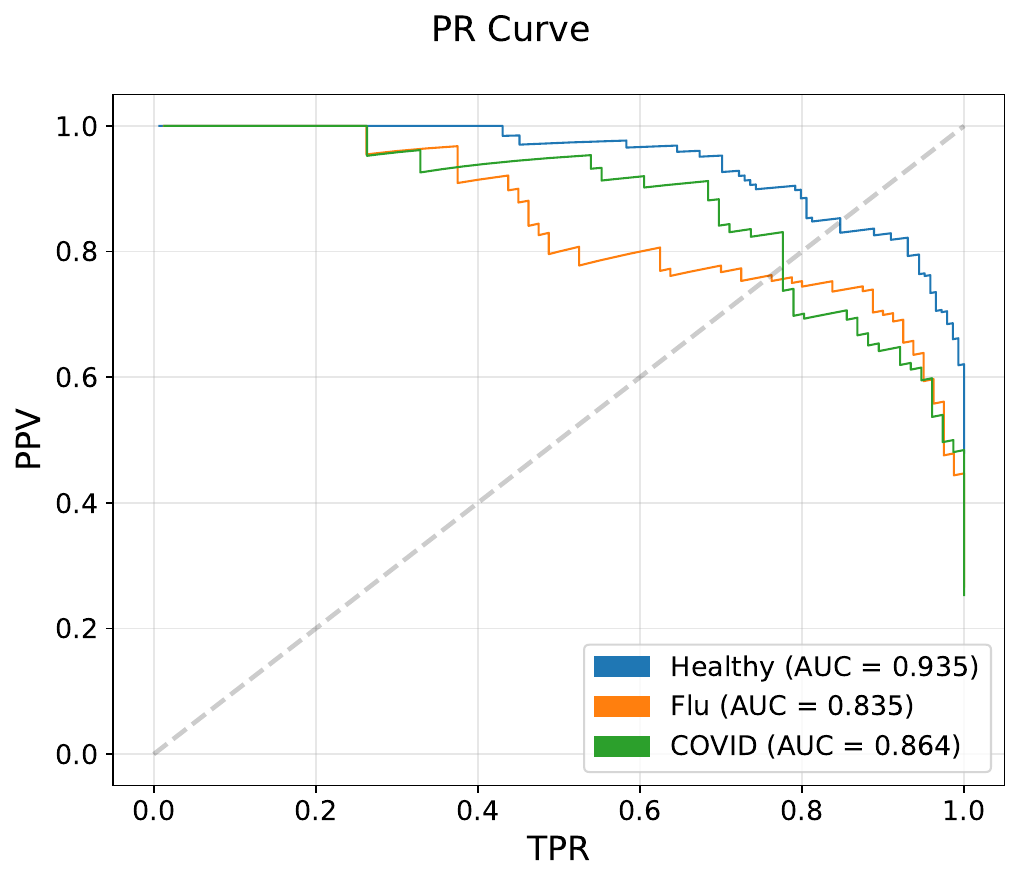}
        \caption{}
        \label{fig:curves:pr}
    \end{subfigure}

    \caption{ROC curves (a) and PR curves (b) for a multi-class classifier distinguishing between Healthy, Flu, and COVID cases as our running example for a hypothetical classifier. Area under the curve (AUC) values are displayed for each class.
    These values highlight how the model’s performance varies across the different categories.}
    \label{fig:curves}
\end{figure}

\textbf{ROC (Receiver Operating Characteristic)} plots the True Positive Rate (TPR) against the False Positive Rate (FPR), defined as $\text{Curve}_{TPR, FPR}$ as a graphical representation of binary
classifiers’ performance~\cite{fawcett2006introduction}. Figure~\ref{fig:curves:roc} illustrates an example ROC curve. An ideal classifier corresponds to the top-left corner of the plot, and a larger Area Under the Curve (AUC) generally indicates superior discriminative performance. In \pycm, this is implemented via the \fbox{\Verb|ROCCurve|} class, which supports both user-defined and automatically generated thresholds. 

\textbf{PRCurve (Precision-Recall curve)} plots Precision (PPV) against Recall (TPR), defined as $\text{Curve}_{PPV, TPR}$.
Figure~\ref{fig:curves:pr} illustrates an example PR curve.
For this visualization, the ideal performance point is the top-right corner, and a larger area under the curve represents better performance~\cite{fawcett2006introduction}.
In \pycm, this is implemented via the \fbox{\Verb|PRCurve|} class, which supports both user-defined and automatically generated thresholds.

\section{Comparing Multiple Classifiers}
\label{sec:compare}
Evaluating and comparing classification models becomes increasingly challenging as the number of classes grows or when the available metrics fail to fully capture performance. Individual evaluation metrics often provide only a partial view of model quality, and manually examining multiple metrics simultaneously can lead to inconsistent or subjective conclusions. In many practical scenarios, it is also necessary to tune an algorithm by systematically examining how changes in its hyperparameters affect the resulting confusion matrix. These challenges underscore the need for a unified approach to compare confusion matrices in a structured, reproducible way.

The \fbox{\Verb|Compare|} class in \pycm addresses this need by providing a benchmarking method that aggregates several existing evaluation criteria into two composite scores. Instead of forcing the user to examine many separate metrics, the method summarizes each model's performance using a combination of overall and class-based benchmarks. Each benchmark evaluates performance on a scale from 0 to 1, where 1 indicates good performance, and 0 represents poor performance.

For each confusion matrix, two scores are calculated. The overall score is the average of seven well-established overall benchmarks. These are Landis and Koch~\cite{landis1977measurement}, Cramer~\cite{lee2016alternatives}, Matthews~\cite{hinkle2003applied}, Goodman-Kruskal Lambda A~\cite{villela2014survey}, Goodman-Kruskal Lambda B~\cite{villela2014survey}, Krippendorff Alpha~\cite{saura2019black}, and Pearson C~\cite{schubert2007importance}.
The class-based score is the average of six class-level benchmarks. These are the Positive Likelihood Ratio Interpretation~\cite{bekkar2013evaluation}, Negative Likelihood Ratio Interpretation~\cite{raslich2007selecting}, Discriminant Power Interpretation~\cite{bekkar2013evaluation}, AUC Value Interpretation~\cite{bekkar2013evaluation}, Matthews Correlation Coefficient Interpretation~\cite{hinkle2003applied}, and Yule Q Interpretation~\cite{bohrnstedt1994statistics}. If any benchmark returns a value of none for a particular class, that benchmark is removed from the averaging. If the user specifies class weights, the class-based score becomes a weighted average rather than a simple average. The scoring formula is presented below.
\begin{align}
    \label{compare}
    S_{\text{Overall}}&=\sum_{i=1}^{|B_O|}\frac{W_{OB}(i)}{\sum W_{OB}}\times\frac{R_O(i)}{|B_O(i)|},\nonumber\\
    S_{\text{Class}}&=\sum_{i=1}^{|B_C|}\sum_{j=1}^{|C|}\frac{W_{CB}(i)}{\sum W_{CB}}\times\frac{W_C(j)}{\sum W_{C}}\times\frac{R_C(i,j)}{|B_C(i)|},
\end{align}
where,
$B_C$ represents class benchmarks,
$B_O$ shows overall benchmarks,
$C$ shows classes,
$W_{CB}$ is class benchmark weights,
$W_{OB}$ is overall benchmark weights,
$W_{C}$ is class weights,
$R_C$ is the class benchmark result,
$R_O$ is the overall benchmark result,
and $|\cdot|$ is the cardinality of a benchmark, which is its maximum possible score.

Once the composite scores are obtained, the method identifies the best confusion matrix based on user preferences. If the parameter \texttt{by\_class} is set to \texttt{True}, the comparison selects the matrix with the highest class-based score. Otherwise, a matrix must achieve the maximum value for both overall and class-based scores in order to be identified as the best. In situations where no matrix simultaneously maximizes both, no best matrix is reported. This rule ensures that both dimensions of performance are considered unless the user explicitly prioritizes class-level behavior.

Therefore, \pycm \fbox{\Verb|Compare|} simplifies the process of model comparison, provides clear insight into model quality, and enables hyperparameter optimization based directly on confusion matrix performance. It reduces the need for manual examination of multiple metrics and helps standardize decision-making when evaluating many models or parameter sets.

It is important to note that this comparison is not an absolute measure of superiority and should be interpreted as a suggestion and not a definitive judgment. The relative importance of different benchmarks can vary across applications. Although the method has been used effectively in several cases~\cite{sannasi2023multi,hanson2022diversity,prasad2022forecasting}, users should consider the characteristics of their specific problem when interpreting the results.

\section{Case Study}
\label{sec:case_study}
To illustrate that comparing models' performance can be non-trivial, we present a practical application of \pycm using the Covertype dataset~\cite{covertype_31}. The dataset contains approximately 600k samples described by 54 cartographic variables (both categorical and integer), such as slope and soil type. The objective is to predict forest cover types using these 54 features. The dataset is accessible via the \texttt{sklearn}\footnote{https://scikit-learn.org/} library, and a standard train/test splitting is established as follows:
\pagebreak
\begin{python}
from sklearn.datasets import fetch_covtype
from sklearn.model_selection import train_test_split

# Data loading
cov_type = fetch_covtype()
X = cov_type.data
y = cov_type.target

# Data Splitting
X_train, X_test, y_train, y_test = train_test_split(
    X, y, test_size=0.4, random_state=0, stratify=y)
\end{python}
For comparative evaluation, we initialize two decision tree classifiers with identical hyperparameters, differing only in the \texttt{random\_state} value. After training, both models produce predictions of the cover type on the test set.

\begin{python}
from sklearn.tree import DecisionTreeClassifier

# Training
dt1 = DecisionTreeClassifier(splitter="random", random_state=1).fit(X_train, y_train)
dt2 = DecisionTreeClassifier(splitter="random", random_state=3).fit(X_train, y_train)

# Inference
pred1 = dt1.predict(X_test)
pred2 = dt2.predict(X_test)
\end{python}

The performance of these two classifiers is then evaluated using \pycm based on different class-based metrics (such as accuracy and f1-score), their averages (macro and micro), and also their respective confusion matrix. The following implementation demonstrates the extraction of these statistics and the generation of the confusion matrix.
\begin{python}
# Evaluation
from pycm import ConfusionMatrix

# Evaluation of the first classifier (can be replicated for cm2)
cm1 = ConfusionMatrix(actual_vector=y_test, predict_vector=pred1)
print('ACC:', cm1.ACC) # class-based accuracy
print('ACC_MACRO:', cm1.ACC_Macro) # Macro averaged accuracy
print('F1:', cm1.F1) # class-based f1-score
print('F1_MACRO:', cm1.F1_Macro) # Macro averaged f1-score
cm1.print_matrix() # confusion matrix
\end{python}

\begin{table}[htbp]
\centering
\begin{tabular}{lcccc}
\toprule
\multirow{2}{*}{\textbf{Class}} & 
\multicolumn{2}{c}{\textbf{Classifier 1}} & 
\multicolumn{2}{c}{\textbf{Classifier 2}} \\
\cmidrule(lr){2-3} \cmidrule(lr){4-5}
& \textbf{Accuracy} & \textbf{F1-Score} & \textbf{Accuracy} & \textbf{F1-Score} \\
\midrule
1: Spruce/Fir          & 0.9433 & 0.9221 & \textbf{0.9455} & \textbf{0.9253} \\
2: Lodgepole Pine      & 0.9370 & 0.9354 & \textbf{0.9394} & \textbf{0.9378} \\
3: Ponderosa Pine      & 0.9895 & 0.9147 & \textbf{0.9900} & \textbf{0.9188} \\
4: Cottonwood/Willow   & \textbf{0.9984} & \textbf{0.8242} & 0.9981 & 0.7949 \\
5: Aspen               & 0.9933 & 0.7957 & \textbf{0.9937} & \textbf{0.8068} \\
6: Douglas-fir         & 0.9911 & 0.8509 & \textbf{0.9917} & \textbf{0.8614} \\
7: Krummholz           & \textbf{0.9955} & \textbf{0.9357} & \textbf{0.9955} & 0.9354 \\
\midrule
Macro Average & 0.9783 & 0.8827 & \textbf{0.9791} & \textbf{0.8829} \\
\bottomrule
\end{tabular}
\caption{Classifiers Performance Summary (Per-class Accuracy and F1-score). Boldface values denote the higher metric's value within each class. Overall, Classifier 2 achieved slightly better performance than Classifier 1, with higher scores in most classes and in the macro averaged metrics.}
\label{tab:classifier-performance-summary}
\end{table}
\begin{figure}[htbp]
    \centering
    \begin{subfigure}{0.48\textwidth}
        \centering
        \includegraphics[width=\linewidth]{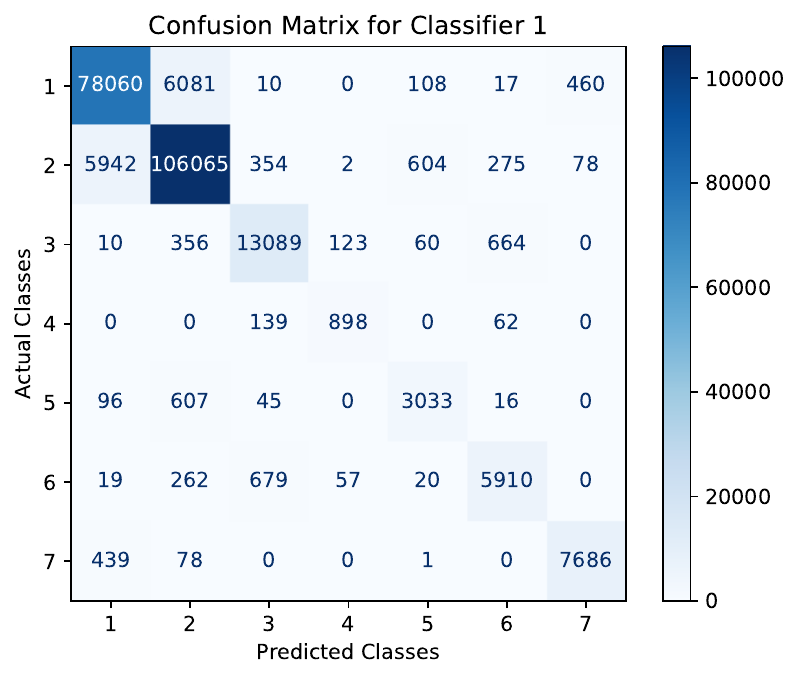}
        \caption{}
        \label{fig:confusion_matrix_classifier:cm1}
    \end{subfigure}
    \hfill
    \begin{subfigure}{0.48\textwidth}
        \centering
        \includegraphics[width=\linewidth]{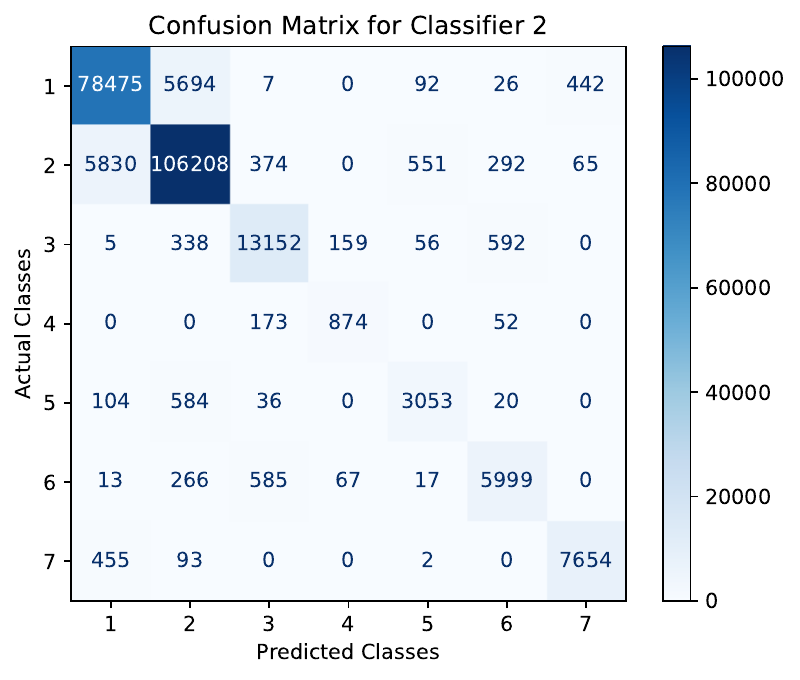}
        \caption{}
        \label{fig:confusion_matrix_classifier:cm2}
    \end{subfigure}

    \caption{Comparison of confusion matrices for (a) Classifier 1 and (b) Classifier 2. Each matrix shows the number of predictions per class.}
    \label{fig:confusion_matrix_classifier}
\end{figure}

Table \ref{tab:classifier-performance-summary} summarizes the accuracy and F1-scores for both classifiers, while Figure \ref{fig:confusion_matrix_classifier} provides the respective confusion matrices. Based on these aggregate metrics, classifier 2 appears to be the stronger candidate for general deployment. However, such high-level summaries can obscure important weaknesses in specialized or high-stakes settings.

To examine the importance of the context, here we present two evaluation scenarios. The first scenario focuses on identifying forest types with high flammability, such as Lodgepole Pine, where accurate detection is vital for wildfire risk assessment due to the presence of serotinous cones~\cite{lotan1985role}. The second scenario addresses the identification of riparian forest types, such as Cottonwood/Willow, which serve as essential ecological indicators for water-adjacent habitats~\cite{stromberg1993fremont}. In both cases, \pycm is utilized to provide a granular assessment of how effectively each model performs on these specific, priority categories.

\begin{table}[htb]
\def\arraystretch{1.1}
\centering

\begin{tabular}{l|cc}
\toprule
\textbf{Class} & \textbf{Flammability} & \textbf{Riparian} \\
\hline
Spruce/Fir       & 0.4 & 0.1 \\
Lodgepole Pine   & 0.9 & 0.0 \\
Ponderosa Pine   & 0.6 & 0.0 \\
Cottonwood/Willow & 0.1 & 0.9 \\
Aspen            & 0.3 & 0.2 \\
Douglas-fir      & 0.7 & 0.1 \\
Krummholz        & 0.5 & 0.0 \\
\bottomrule
\end{tabular}
    \caption{Relative flammability and riparian scores for forest cover types in the Covertype dataset. These values are nominal, intended solely for illustrative purposes in this example.}
\label{tab:flam-rip}
\end{table}

\subsection{Flammability Assessment}
In wildfire management, decision-making must prioritize the correct identification of highly flammable land cover. To reflect this requirement, we assign each of the seven cover types a flammability index ranging from 0 to 1 (Table \ref{tab:flam-rip}), where larger values indicate greater fire susceptibility. These indices are interpreted as importance weights that represent the relative cost of misclassification for each class. By incorporating these weights directly into the \fbox{\Verb|Compare|} module, model performance can be evaluated in a manner that accounts for fire risk rather than treating all classes as equally important:
\pagebreak

\begin{python}
from pycm import Compare

flammability_score = {
    1: 0.4,   # Spruce/Fir 
    2: 0.9,   # Lodgepole Pine
    3: 0.6,   # Ponderosa Pine
    4: 0.1,   # Cottonwood/Willow
    5: 0.3,   # Aspen
    6: 0.7,   # Douglas-fir
    7: 0.5,   # Krummholz
}
compare = Compare({'Classifier 1': cm1, 'Classifier 2': cm2}, class_weight=flammability_score)
print(compare)
# Best : Classifier 2
# 
# Rank  Name            Class-Score       Overall-Score
# 1     Classifier 2    0.88452           0.92381
# 2     Classifier 1    0.88143           0.92381
\end{python}

Aligned with the overall performance metrics, the \fbox{\Verb|Compare|} module also ranks Classifier 2 as the stronger predictor for this task. In contrast to conventional evaluations, which involve multiple metrics and potentially conflicting interpretations, the \fbox{\Verb|Compare|} module consolidates performance into a single comparative ranking, thereby simplifying the model selection process.

\subsection{Riparian Zone Detection}
Accurate identification of riparian zones, the land areas adjacent to rivers that support distinct vegetation patterns and water-dependent habitats, is essential in environmental monitoring. To determine which of the previously trained models is better suited for this task, we evaluate their performance using importance weights derived from vegetation likelihoods. Table \ref{tab:flam-rip} lists the relative probability of each cover type occurring near a river.

In this setting, these probabilities are treated as class-importance weights, guiding the evaluation to emphasize cover types that are more representative of riparian environments. Incorporating these weights shifts the comparison from overall accuracy toward the models’ reliability in identifying water-adjacent ecosystems.

\begin{python}
from pycm import Compare

riparian_score = {
    1: 0.1,   # Spruce/Fir
    2: 0.0,   # Lodgepole Pine
    3: 0.0,   # Ponderosa Pine
    4: 0.9,   # Cottonwood / Willow
    5: 0.2,   # Aspen
    6: 0.1,   # Douglas-fir
    7: 0.0,   # Krummholz
}
compare = Compare({'Classifier 1': cm1, 'Classifier 2': cm2}, class_weight=riparian_score)
print(compare)
# Best : Classifier 1
# 
# Rank  Name            Class-Score       Overall-Score
# 1     Classifier 1    0.86218           0.92381
# 2     Classifier 2    0.79295           0.92381
\end{python}

In contrast to the overall performance analysis, which favored Classifier 2, the \fbox{\Verb|Compare|} module identifies Classifier 1 as the more suitable model for this task. Although its overall performance is lower, it achieves better results on the classes most relevant to riparian zone detection.

This change in ranking underscores a key limitation of relying solely on global performance metrics: optimizing for average accuracy can lead to suboptimal choices in task-specific settings. By incorporating class-importance weights that reflect the application's priorities, the \texttt{Compare} module aligns the evaluation with the practical objective and supports selecting the model that performs best where it is most needed.

\section{Related Work}
This tutorial bridges the gap between existing computational tools and the pedagogical literature surrounding model assessment. By pairing hands-on examples with a robust theoretical foundation, we hope to transform the model selection process into a truly informed decision-making exercise.

\subsection{Comparative Software Frameworks}
The Python ecosystem offers several libraries for performance measurement, most notably Scikit-learn~\cite{pedregosa2011scikit}. While Scikit-learn provides a robust suite of basic metrics through its metrics module, it is primarily designed as a general-purpose modeling toolkit. Consequently, it lacks the specialized, exhaustive statistical metrics, such as the G-measure or multi-class generalizations of the Matthews Correlation Coefficient (MCC), that \pycm provides natively.

Outside of Python, the Caret package in R~\cite{kuhn2008building} has long been the standard for comprehensive confusion matrix analysis. This library offers built-in statistics like Cohen’s Kappa and McNemar’s Test. However, for researchers working within a Pythonic workflow, a gap exists for a library that matches Caret’s statistical breadth while offering seamless integration with modern data science stacks.
Other specialized tools like MLxtend~\cite{raschka2018mlxtend} focus heavily on developing machine learning algorithms that has been left out by other libraries but do not reach the same depth of descriptive statistical reporting required for high-stakes model auditing.

\subsection{Theoretical and Pedagogical Literature}
A significant body of work exists regarding the theory of classification error. Many tutorials focus on the Binary Classification case, neglecting the nuances of Multi-class dimensionality~\cite{sokolova2009systematic,fawcett2006introduction}. Existing educational resources treat evaluation as a secondary step to model training, often providing the ``how'' of API calls without the ``why'' of metric selection.
Answering to this why questions become even more important when building a multi-class framework. 
This tutorial builds upon the foundations laid by \citet{haghighi2018pycm}, who introduced PyCM as a solution for researchers needing a deep dive into model behavior. By focusing on case-specific scenarios, this tutorial extends the existing literature by providing a practical roadmap for interpreting conflicting metrics in complex datasets.

\subsection{Emerging Trends in Model Auditing}
Recent shifts in the field of model auditing have moved toward algorithmic fairness~\cite{kim2019multiaccuracy,yang2024large}, where a single score is no longer considered sufficient for validation~\cite{barocas-hardt-narayanan}. This has led to the development of frameworks that emphasize per-class performance and the stability of metrics across different data distributions. This tutorial presents \pycm's usability in this trend by providing a unified interface that allows for the simultaneous calculation of over 40 metrics, enabling a multi-dimensional perspective.
\section{Conclusion}
Effective model evaluation requires looking beyond simple metrics like accuracy. Through two case studies, we demonstrated that relying on a single metric can hide critical performance details and trade-offs. We illustrated how the \pycm library allows practitioners to efficiently compute, compare, and interpret a wide range of metrics, providing a deeper understanding of model behavior.

This tutorial serves as a guide to select appropriate evaluation methods and integrate rigorous analysis into the machine learning workflow. Ultimately, note that a comprehensive evaluation strategy ensures models reliability and accuracy in real-world applications.
\section*{Acknowledgments}
We acknowledge that part of this project was funded by the NGI0 Commons Fund, part of the European Commission's Next Generation Internet programme (Grant No. 101135429).

\bibliography{_refrences}
\bibliographystyle{iclr2025_conference}


\end{document}